\documentclass[sigconf]{acmart}

\usepackage{listings}
\usepackage{xcolor}

\definecolor{codebackground}{rgb}{0.95,0.95,0.95}
\definecolor{codekeyword}{rgb}{0.2,0.4,0.6}
\definecolor{codecomment}{rgb}{0.4,0.4,0.4}

\lstdefinestyle{promptstyle}{
    backgroundcolor=\color{codebackground},
    basicstyle=\small\ttfamily,
    breaklines=true,
    captionpos=b,
    commentstyle=\color{codecomment},
    keywordstyle=\color{codekeyword},
    stringstyle=\color{blue},
    numbers=none,
    numbersep=5pt,
    showspaces=false,
    showstringspaces=false,
    showtabs=false,
    tabsize=2,
    frame=single,
    framesep=5pt,
    framerule=0.4pt,
    xleftmargin=8pt,
    xrightmargin=8pt,
    aboveskip=8pt,
    belowskip=8pt
}
\AtBeginDocument{%
  }

\setcopyright{acmlicensed}
\copyrightyear{2025}

\acmYear{2025}

\setcopyright{rightsretained}

\acmConference[KDD '25] {Proceedings of the 1st Workshop on "AI for Supply Chain: Today and Future" @ 31st ACM SIGKDD Conference on Knowledge Discovery and Data Mining V.2}{August 3, 2025}{Toronto, ON, Canada.}

\acmBooktitle{Proceedings of the 1st Workshop on "AI for Supply Chain: Today and Future" @ 31st ACM SIGKDD Conference on Knowledge Discovery and Data Mining V.2 (KDD '25), August 3, 2025, Toronto, ON, Canada}

\acmISBN{979-8-4007-1454-2/25/08}

\acmDOI{10.1145/XXXXXX.XXXXXX}

\begin{document}

% \title{Multi-Agent Automation for Knowledge Base Construction from Unstructured Supply Chain Data}
% From Unstructured Communication to Intelligent RAG: Multi-Agent Automation for Supply Chain Knowledge Bases
\title{From Unstructured Communication to Intelligent RAG: Multi-Agent Automation for Supply Chain Knowledge Bases}

\author{Yao Zhang}
%% \authornote{Both authors contributed equally to this research.}
\email{yaozhanq@amazon.com}
\orcid{0000-0001-8096-7112}
\affiliation{%
  \institution{Amazon Operational Technology Solutions}
  \city{Austin}
  \state{Texas}
  \country{USA}
}

\author{Zaixi Shang}
\orcid{0000-0002-4264-3130}
\email{zaishang@amazon.com}
\affiliation{%
  \institution{Amazon Operational Technology Solutions}
  \city{Bellevue}
  \state{WA}
  \country{USA}
}

\author{Silpan Patel}
\email{silpan@amazon.com}
\affiliation{%
  \institution{Amazon Operational Technology Solutions}
  \city{Austin}
  \state{Texas}
  \country{USA}
}

\author{Mikel Zuniga}
\email{mkzng@amazon.com}
\affiliation{%
  \institution{Amazon Operational Technology Solutions}
  \city{Seattle}
  \state{WA}
  \country{USA}
}

\renewcommand{\shortauthors}{Zhang et al.}

\begin{abstract}
Supply chain operations generate vast amounts of operational data; however, critical knowledge—such as system usage practices, troubleshooting workflows, and resolution techniques—often remains buried within unstructured communications like support tickets, emails, and chat logs. While Retrieval-Augmented Generation (RAG) systems aim to leverage such communications as a knowledge base, their effectiveness is limited by raw data challenges: support tickets and chat logs are typically noisy, inconsistent, and incomplete, making direct retrieval suboptimal. Unlike existing RAG approaches that focus on runtime optimization, we introduce a novel \textit{offline-first} methodology that transforms these communications into a structured knowledge base. Our key innovation is a Large language models (LLMs)-based multi-agent system orchestrating three specialized agents: Category Discovery for taxonomy creation, Categorization for ticket grouping, and Knowledge Synthesis for article generation. Applying our methodology to real-world support tickets with resolution notes and comments between resolvers and requesters, our system creates a compact knowledge base—reducing the total volume to just 3.4\% of the original ticket data while improving quality. Experiments demonstrate that plugging our prebuilt knowledge base into a RAG system significantly outperforms traditional RAG implementations (48.74\% vs. 38.60\% helpful answers) and achieves a 77.4\% reduction in unhelpful responses. By automating institutional knowledge capture that typically remains siloed in experts' heads, our solution translates to substantial operational efficiency: reducing support workload, accelerating resolution times, and creating self-improving systems that can automatically resolve approximately 50\% of future supply chain tickets. Our approach addresses a key gap in knowledge management by transforming transient communications into structured and reusable knowledge base through intelligent offline processing rather than latency-inducing runtime architectures.
\end{abstract}





\maketitle

\section{Introduction}

% - Supply chain complexity and knowledge challenges
% Knowledge management improves supply chain performance. 
Modern supply chains generate enormous data volumes, yet operational knowledge—how to navigate order systems, track shipments, troubleshoot fulfillment errors, or debug inventory mismatches—remains buried in unstructured communications like emails, tickets, incident reports, and chat logs. Supply chain knowledge presents management challenges due to its distribution across organizational silos, mixture of structured data and unstructured communications, temporal evolution, and combination of explicit and tacit components. Ticket systems, where operations are documented and exceptions are resolved, become repositories of valuable procedures, problem-solving approaches, workarounds, and domain expertise. However, this knowledge typically remains isolated in communication threads, causing solutions to be rediscovered rather than reused.

% - Limitations of current approaches
Traditional knowledge management approaches in supply chain rely heavily on manual documentation processes that are time-consuming and inconsistent. Static knowledge bases quickly become outdated in the dynamic supply chain environment. Moreover, these approaches struggle to capture the context and nuance from communications where much of the problem-solving occurs. Particularly challenging is the capture of institutional knowledge—the unwritten information, processes, and expertise known by experienced team members but not formally documented. 

% - Overview of our automated knowledge base creation solution
To automate knowledge management to improve supply chain performance, we present a Large language models (LLMs)-based multi-agent system designed specifically for extracting, organizing, and utilizing supply chain knowledge from ticket systems. Our approach transforms unstructured communications in ticket systems into a structured knowledge base through three specialized agents working in a coordinated pipeline:

\begin{enumerate}
    \item \textbf{Category Discovery Agent}: Analyzes ticket data to identify distinct knowledge domains and creates a taxonomy of categories that forms the organizational structure of the knowledge base.
    \item \textbf{Ticket Categorization Agent}: Assigns each ticket to one or more relevant categories, grouping related issues to enable effective knowledge synthesis.
    \item \textbf{Knowledge Synthesis Agent}: Transforms groups of categorized tickets into comprehensive knowledge articles that capture generalizable insights, common patterns, and proven solutions.
\end{enumerate}

The system operates end-to-end for knowledge base creation, with the Category Discovery Agent first creating an initial taxonomy, which the Categorization Agent then uses to classify tickets. The Knowledge Synthesis Agent processes these categorized tickets to generate structured knowledge articles, with feedback mechanisms allowing refinements to the category structure through subcategory discovery and specialized synthesis when needed. This approach addresses key challenges in supply chain knowledge management by automatically transforming transient communications into reusable and structured knowledge. It captures institutional knowledge that would otherwise remain siloed, standardizes terminology and problem-solving approaches, and creates a foundation for automated support systems. The resulting knowledge base serves as the context source for a Retrieval-Augmented Generation (RAG) system, enabling more effective automated responses to supply chain queries.

% - Key Contributions
Our work makes the following contributions:

\begin{itemize}
    \item We propose a simple yet effective framework for multi-agent knowledge base creation from communication data in ticket systems that minimizes implementation complexity via LLM calls while maximizing knowledge synthesis quality, addressing the challenge of knowledge organization at scale.
    
    \item We evaluate our approach on real-world ticket communications and demonstrate that integrating the resulting knowledge base with a RAG model significantly improves the helpfulness of generated answers, assisting in the resolution of about 50\% of future tickets.
    
    \item We establish an offline-first methodology for RAG enhancement that complements existing runtime optimization approaches. While most related work focuses on improving RAG systems online during query processing with scalability concern: chaining LLM calls (multiple agents or iterative retrieval) can incur high latency and cost, we demonstrate that substantial gains can be achieved through offline knowledge base creation, which can also be combined with online improvements at runtime.
\end{itemize}

Our work builds upon established theoretical frameworks in knowledge management, particularly Nonaka and Takeuchi's SECI model of knowledge creation and transformation \cite{nonaka1995}, by providing an automated mechanism for converting tacit knowledge (embedded in communication) to explicit knowledge (structured knowledge articles).

\section{Related Work}

\subsection{LLM-Driven Knowledge Extraction and Summarization}
% LLMs can extract entities/relations and summarize content to form knowledge graphs or KBs. They excel at semantic enrichment when combined with graph structures. These methods tend to outperform standalone LLMs or static retrieval on comprehension and coverage. A limitation is that LLMs may still hallucinate without grounding, so many systems track provenance or use multi-stage pipelines for reliability.

Enterprise knowledge extraction has recently leveraged LLMs to parse and synthesize unstructured text (tickets, chat logs, documents) into structured insights. A summary of related work is in the Appendix. Table \ref{tab:llm-research}. For example, Anderson et al. \cite{anderson_design_2025} introduce an LLM-powered analytics system that parses documents into "DocSets" and uses LLMs for semantic query planning and document operations. In another line of work, Kumar et al. \cite{kumar_llm-powered_2025} propose an LLM-driven activity-centric knowledge graph for enterprises. Their framework ingests emails, calendars, chat logs and documents, uses LLMs to extract entities/relations and enrich semantics, and builds a unified knowledge graph (KG), which reportedly improves workflows like meeting preparation and analytics-driven decision-making.
Other works focus on summarization as a vehicle for knowledge capture. Edge et al. \cite{edge_local_2025} propose GraphRAG: an LLM-based workflow that first builds an entity-centric knowledge graph from a corpus and pre-computes "community summaries" for clusters of related entities. At query time, each cluster's summary is generated and combined into a final answer. This multi-stage approach improves coverage and diversity on long-text summarization tasks compared to vanilla RAG. 
Also, Li et al. \cite{li_efficient_2025} introduced EDC²-RAG, a dynamic clustering-based document compression framework that enhances RAG performance by reducing redundancy and noise in retrieved documents through latent inter-document relationship modeling and query-specific summarization. 
Similarly, in support and IT service domains, several works have applied LLMs to summarize and structure ticket data. Isaza et al.~\cite{isaza_retrieval_2024} present a pipeline that processes IT support tickets through classification, query generation, and solution summarization using fine-tuned LLMs. Wulf and Meierhofer~\cite{wulf_exploring_2024} explore large-scale automation of technical support ticket handling via RAG-based summarization, while Sun et al.~\cite{sun_llm-augmented_2025} propose LLM-driven aggregation and clustering of mobile OS defect tickets to enhance resolution efficiency.
More generally, LLMs have been used to extract knowledge graphs from text. For example, Zhang et al. \cite{zhang_llms_2024} present an LLM-instructed KG approach: an LLM reads complex text to build a knowledge graph, then edits a domain-specific LLM with that graph. This two-stage "Sequential Fusion" improved domain-LLM QA accuracy to ~72–75\% on medical and economics QA tasks. Such LLM-driven KG construction and summarization pipelines illustrate how modern LLMs can synthesize structured knowledge (entities, relations, summaries) from raw text.
Unlike these approaches that focus on entity extraction and relationship mapping, our work emphasizes hierarchical knowledge organization through categorical taxonomy creation and multi-level synthesis.

% \subsection{Case Studies in Supply Chain Domains}
% % In supply chains, LLMs enable automated news monitoring and supplier-network mapping. In general enterprise, LLM pipelines can auto-build knowledge graphs from emails/chats or automate document workflows like contract review. While results are often promising, they usually rely on careful pipeline design and still require human oversight.

% % These methods for LLM-Driven knowledge extraction and summarization, RAG, and multi-agent system have been applied in both general enterprise and specific domains like supply chain. In the supply chain domain, several works highlight LLM-driven KB creation: 
In supply-chain contexts, Houamegni and Gedikli \cite{houamegni_evaluating_2025} evaluate LLMs (GPT-4, Mistral, etc.) on summarizing news for supply-chain risk analysis. They find that modern LLMs (especially a few-shot GPT-4o mini) produce much higher-quality summaries (by ROUGE/BLEU and human judgment) than older models, effectively highlighting relevant risk incidents. 
% Houamegni and Gedikli \cite{houamegni_evaluating_2025} present an LLM-based news summarizer for supply-chain risk. They aggregate news feeds and prompt LLMs to produce concise risk digests for companies. Their evaluation (ROUGE/BLEU and user study) shows top LLMs (GPT-4o) significantly boost summary quality and risk identification.
Likewise, AlMahri et al. \cite{almahri_enhancing_2024} use LLMs (zero-shot prompting) to extract supply-chain relationships from public data into a knowledge graph. Focusing on electric-vehicle batteries, they automatically build a multi-tier supplier graph (identifying mineral sources, manufacturers, etc.) without needing partner disclosures. Their case study shows that the LLM-derived graph improves visibility beyond first-tier suppliers, aiding risk management and strategic sourcing. 
While these approaches apply LLMs to supply chain data, they focus on external information sources (news, public data) rather than internal operational communications. Our work specifically addresses the challenge of transforming internal operational knowledge from support tickets into a structured knowledge base, capturing the valuable institutional knowledge that remains hidden in organizational communications.
\subsection{Retrieval-Augmented Generation (RAG) in Enterprise Contexts}
% RAG is effective for factual QA, but needs enhancement for complex enterprise tasks. Techniques like graph-based summarization or interleaved retrieval add multi-hop reasoning. Architectures like ADW explicitly blend RAG with parsing and procedural steps. A limitation is system complexity – each extra stage or agent adds engineering overhead and potential latency.

RAG, retrieving relevant text chunks as LLM context, is now common for enterprise QA and summarization. Recent work has explored enterprise-scale deployments and operational considerations: RAG at Scale: Lessons from Building an Enterprise QA System \cite{alcaraz2024enterprise} examines architecture and performance trade-offs when indexing millions of documents; SecureRAG: Privacy-Preserving Retrieval-Augmented Generation in Enterprises \cite{polymersolutions2025securerag} introduces encrypted indexes and access control to protect sensitive corporate data; and RAGOps: Operationalizing Retrieval-Augmented Generation for Large-Scale Customer Support \cite{kimothi2024ragops} presents an end-to-end pipeline for data ingestion, real-time indexing, monitoring, and disaster recovery in production RAG services. 
While these works focus primarily on the operational aspects of RAG deployment, our approach addresses the fundamental issue of knowledge quality by transforming raw data into a structured knowledge base before it enters the RAG pipeline.

Studies note that naïve RAG is often insufficient for complex tasks. Anderson et al. \cite{anderson_design_2025} point out that simple "hunt-and-peck" RAG handles factoid queries well, but fails on "sweep-and-harvest" problems that require scanning large corpora and synthesizing answers. In these cases, the answer spans many documents or requires aggregation (e.g., "yearly revenue growth of companies with CEO change"). RAG's limitation is that it only pulls a few chunks into the prompt, so it cannot naturally perform long-range reasoning or aggregation. To address this, several works augment RAG with additional structure or steps. 
LlamaIndex's Agentic Document Workflows (ADW) framework prescribes a multi-stage pipeline: Parse (extract typed data from raw files), Retrieve (RAG-style search with vector indexes), Reason (policy/rule-guided multi-step LLM logic), and Act (output structured results to systems) \cite{llamaindex}. 
% In ADW, RAG is only the "Retrieve" phase – it is combined with parsing and structured synthesis. For example, in a contract risk analysis workflow, ADW first parses clauses via an LLM, then matches clauses against a policy KB, then uses an agent to summarize flagged clauses, and finally connects results to a CLM system. This shows how RAG can feed into a larger agentic pipeline for enterprise automation. 
Another direction is iterative and multi-agent RAG. Pathway describes a "Dynamic Agentic RAG" for legal/financial documents \cite{pathway}. They use separate agents for retrieval and reasoning: the system iteratively alternates between fetching relevant pages (via vector search) and refining the answer (via LLM reasoning). 
% In this "interleaved retrieval and reasoning" scheme, agents can query additional tools (calculators, charts, etc.) and update context dynamically. 
This approach outperforms vanilla RAG on benchmarks (improving MRR and answer quality) by keeping the reasoning grounded in retrieved facts.
In summary, modern RAG systems often combine retrieval with graph structures \cite{edge_local_2025, li_efficient_2025} or iterative workflows to overcome RAG's one-shot limitation.

\subsection{Multi-Agent Systems for Task Automation}
% Multi-agent LLM systems allow complex, multi-step processing. Each agent can use LLM/RAG/tools. Breaking tasks down improves fidelity and enables concurrency. Limitations include system complexity (many LLM calls) and the need for robust agent communication protocols (an active research area).

A recent trend is using LLM-based agents in concert to automate complex tasks. Multi-agent architectures decompose a task into sub-tasks and assign them to specialized LLM “agents”. This mirrors robotic process automation but powered by LLMs. The benefit is twofold: it manages LLM context limits by focusing each agent on a subset of the problem, and it allows explicit reasoning workflows. For example, the BMW Multi-Agent Framework outlines a planner-executor architecture \cite{crawford_bmw_2024}. A Planner agent first decomposes a user request into subtasks. In an example RAG-based QA workflow, the planner breaks a query into simpler questions, and a “BMW Assistant” agent executes them by retrieving information via semantic. The answers are synthesized, then a Verifier agent checks that the final response addresses the original query. This multi-agent pipeline improves reliability: by reasoning and verifying each step, it reduces hallucination. Similarly, EvoPat is a multi-LLM patent analysis agent \cite{wang_evopat_2024}. Instead of one LLM, EvoPat launches several agents (akin to “scientists”), each with a role (e.g. summarizer, innovation spotter, critic). They collectively use RAG (searching patent databases and the web) to gather facts. In human evaluations, EvoPat’s agentic approach produced more informative and accurate patent summaries than a single GPT-4o call. The authors note that breaking tasks into agent-specific jobs yields deeper analysis, especially given LLM token constraints. 
% Agentic workflows are also emphasized in industry-oriented proposals. LlamaIndex’s ADW calls for an Agent Orchestration layer where one can mix deterministic logic with LLM steps \cite{llamaindex}. Pathway’s Dynamic RAG uses separate Supervisor, RAG, and Code/Reasoning agents to manage retrieval and tools \cite{pathway}. 
Across these examples, common patterns emerge: agents may specialize in retrieval (search vectors/KB), reasoning (LLM synthesis), or action (API/tool invocation). This mirrors classical architectures like ReAct or actor-critic loops, but at a larger scale.

% Combining these methods—e.g. using multi-agent workflows to harvest and structure supply-chain documents into an enterprise KG—could enable continual KB creation and reuse across operations. Overall, recent research demonstrates that LLMs, when augmented with RAG and agentic orchestration, can substantially ease the transformation of raw enterprise communications into structured knowledge. However, achieving robust, end-to-end automated pipelines will require addressing open challenges in model accuracy, integration, and human-in-the-loop design.
Multi-agent systems have been applied to knowledge management tasks with increasing sophistication. Chen et al. \cite{chen_survey_2025} and Guo et al. \cite{guo_large_2024} provide comprehensive surveys of LLM-based multi-agent systems that emphasize their potential for complex problem-solving. 
% Despite successes, current approaches have limitations. Multi-stage pipelines can be brittle and require extensive engineering. Evaluation is often narrow (ROUGE, human judgment) and there's little open benchmarking on enterprise corpora. 
% Scalability is another concern: chaining LLM calls (multiple agents or iterative retrieval) can incur high latency and cost. 
For supply-chain knowledge bases, related works suggest promising starts (LLM-based KG extraction from text \cite{almahri_enhancing_2024}, news summarization for risk) \cite{houamegni_evaluating_2025}, but fully automated KB curation remains future work. 

Our multi-agent approach differs from existing systems in several key ways. First, while most multi-agent systems focus on runtime task decomposition and execution, our agents work in a coordinated pipeline specifically designed for knowledge base creation. Second, our architecture is purpose-built for transforming unstructured communications into structured knowledge rather than general problem-solving. Third, our system emphasizes categorization and synthesis rather than the planner-executor or reasoning-verification patterns common in other multi-agent architectures. By specializing each agent for a specific phase of knowledge transformation (category discovery, categorization, and synthesis), we create a more focused and efficient pipeline for knowledge base creation.

\subsection{Synthesis and Research Gap}
Reviewing the literature highlights several key gaps that our approach addresses. First, while numerous works have explored LLM-driven knowledge extraction and RAG techniques, most focus on either entity-relation extraction (building knowledge graphs) or runtime optimization (improving retrieval during query processing). In contrast, our approach emphasizes offline knowledge synthesis and organizational structure through categorical taxonomy. Second, existing supply chain applications of LLMs typically focus on external data sources (news, public reports) rather than internal operational communications, missing the opportunity to capture valuable institutional knowledge embedded in support tickets. Third, current RAG improvements predominantly focus on runtime complexity—adding agents, iterative steps, and complex reasoning during query time—which introduces latency and cost concerns in production environments.

Our offline-first methodology presents a fundamentally different approach to RAG enhancement. Rather than adding complexity during query processing, we perform the complex work of knowledge organization and synthesis as a preprocessing step, enabling more efficient runtime operation. This allows us to apply more sophisticated processing to the knowledge base creation without impacting query response times. Our multi-agent category-driven approach specifically targets the transformation of noisy, transient communications into structured, reusable knowledge—a challenge not adequately addressed by current systems that focus primarily on retrieval mechanisms rather than knowledge quality.

\section{Methodology}
\subsection{Problem Formulation}

Support tickets represent a rich source of operational knowledge in supply chain systems. Each ticket encapsulates not only the problem statement from the requester but also the complete troubleshooting process, resolution notes, and the communication thread between the person requesting support and the resolver addressing the issue. These tickets serve as records of both problems and their solutions, making them valuable sources of operational knowledge. However, this knowledge remains trapped in a format that is not conducive to systematic reuse. Our task is to transform this unstructured ticket data into a structured knowledge base that can be effectively utilized by a RAG system.

Formally, given a dataset of tickets $T = \{t_1, t_2, ..., t_n\}$ where each ticket $t_i$ contains unstructured text (title, description, and resolution comments), our goal is to produce a knowledge base $K = \{k_1, k_2, ..., k_m\}$ where each knowledge article $k_j$ synthesizes generalizable knowledge from multiple related tickets. The ideal knowledge base should capture both standard operating procedures and exception handling scenarios, maximize both coverage (encompassing all significant processes and issues in the dataset) and precision (focusing on actionable, reusable knowledge).

For evaluation, we measure how effectively a RAG system leveraging this knowledge base can answer new supply chain queries. Our primary metric is response helpfulness, rated on a 5-point scale by an independent evaluator.

\subsection{Raw Ticket Indexing Baseline}

Our baseline approach processes raw ticket data with minimal transformation, using the entire ticket content as potential context for the RAG system. This represents a typical "naïve RAG" implementation that directly indexes the cleaned text of tickets (including title, description, and comments) without additional knowledge base creation. The baseline serves as a control to demonstrate the value added by our multi-agent knowledge base creation approach.

\subsection{Multi-Agent Category-Driven Synthesis}

\begin{figure}[t]
\centering
\includegraphics[width=\linewidth]{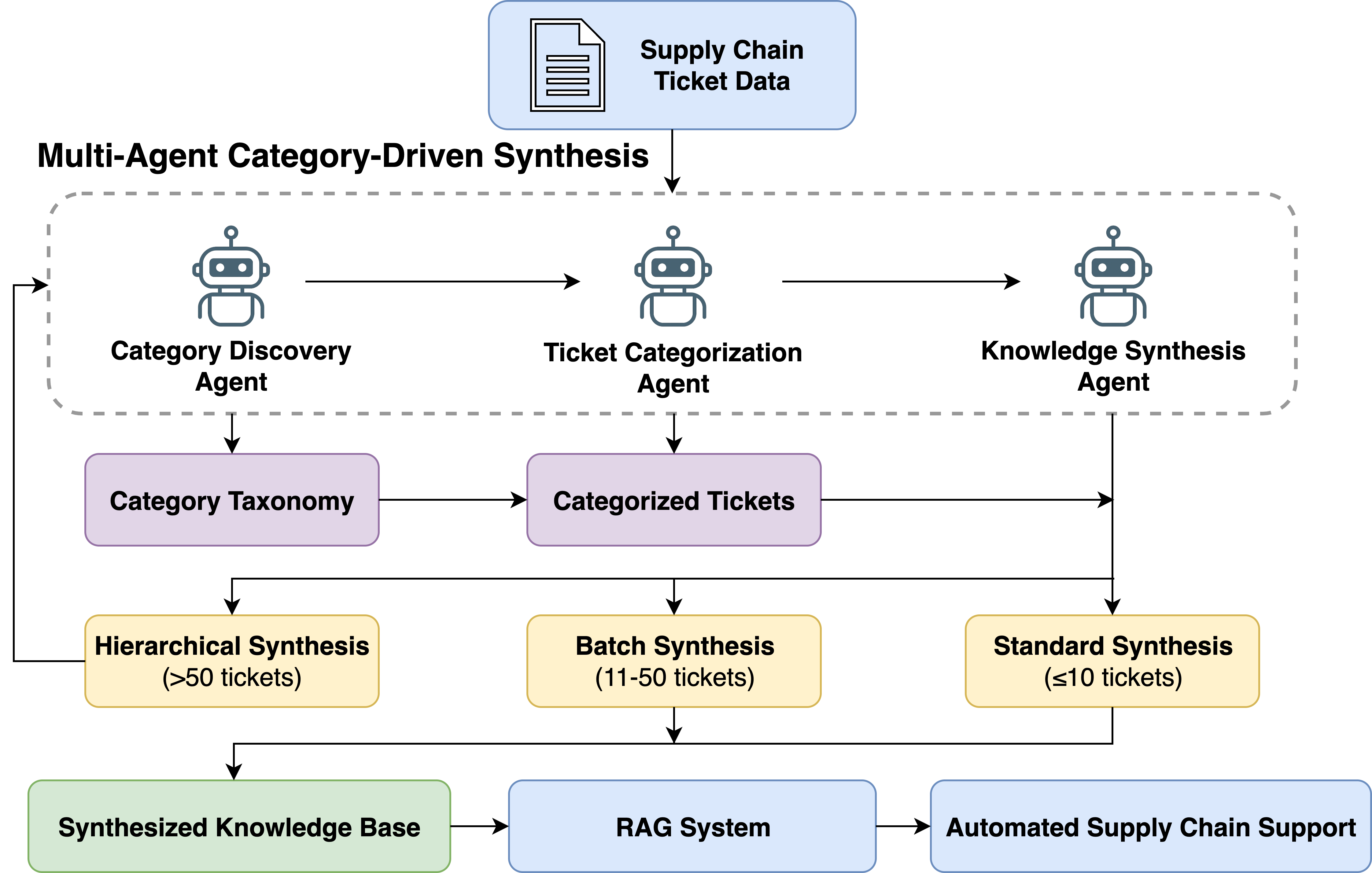}
\caption{Multi-Agent Category-Driven Synthesis Architecture. The system orchestrates three specialized agents working in a coordinated pipeline: Category Discovery Agent identifies knowledge categories from ticket data, Ticket Categorization Agent assigns tickets to appropriate categories, and Knowledge Synthesis Agent transforms categorized tickets into comprehensive knowledge articles using different synthesis strategies based on category size. The resulting synthesized knowledge base serves as the foundation for the RAG system for automated supply chain support.}
\label{fig:multi-agent-architecture}
\end{figure}

We propose a simple yet effective multi-agent approach to knowledge base creation, involving three specialized agents working in a coordinated pipeline shown in Figure \ref{fig:multi-agent-architecture}.
\subsubsection{Category Discovery Agent}
This agent examines a representative sample of tickets to identify knowledge domains that form the organization of our knowledge base. Using Claude Sonnet 3.7 with carefully engineered prompts (see Appendix \ref{category_discovery_prompt}), it analyzes ticket content across the dataset, identifies distinct problem patterns and domains, and creates a taxonomy of categories with clear descriptions and identifying patterns.

We considered two approaches for category discovery: 1. Batch Discovery: Processes tickets in batches independently, then merges results using a specialized merge prompt (see Appendix \ref{category_merge_prompt}); 2. Iterative Discovery: Sequentially processes all tickets, continuously refining categories. While both approaches have merit, we selected Batch Discovery for our experiments due to its superior parallelizability and scalability advantages, which are particularly important for processing large volumes of supply chain tickets.

\subsubsection{Ticket Categorization Agent}
The Categorization Agent assigns each ticket to one or more categories identified by the Discovery Agent. This process groups related issues together for more effective knowledge synthesis. The agent examines each ticket's title and description, compares against all available categories, assigns the ticket to up to two categories that best capture its essence, and provides brief reasoning for each assignment (see Appendix \ref{ticket_categorization_prompt}). We implement parallel processing that significantly accelerates the categorization process.

\subsubsection{Knowledge Synthesis Agent}
The Synthesis Agent transforms groups of categorized tickets into comprehensive knowledge articles. This is the most critical step in our process, as it converts raw ticket data into structured, reusable knowledge. The agent employs several strategies based on category size and complexity, all using the same underlying synthesis prompt (see Appendix \ref{knowledge_synthesis_prompt}):

\begin{enumerate}
    \item \textbf{Standard Synthesis}: For categories with fewer tickets (below a configurable threshold, default 10), processes all tickets in a single pass using the knowledge synthesis prompt.
    
    \item \textbf{Batch Synthesis}: For mid-sized categories, processes different subsets (batches) of tickets independently and in parallel. Each batch uses the same knowledge synthesis prompt to create individual articles, which are then merged into a comprehensive article using the Knowledge Merge prompt (see Appendix \ref{knowledge_merge_prompt}).
\end{enumerate}

For categories with many tickets (exceeding a configurable threshold, default 50), we implement a hierarchical approach:
\begin{enumerate}
    \item \textbf{Subcategory Discovery}: Category Discovery Agent identifies more specific subcategories using a specialized prompt (see Appendix \ref{subcategory_discovery_prompt})
    \item \textbf{Subcategory Categorization}: Ticket Categorization Agent assigns tickets to appropriate subcategories using a specialized prompt (see Appendix \ref{subcategory_categorization_prompt})
    \item \textbf{Subcategory Synthesis}: Knowledge Synthesis Agent creates knowledge articles for each subcategory using the same synthesis prompt as above (see Appendix \ref{knowledge_synthesis_prompt})
    % \item \textbf{Parent Article Creation}: A high-level article provides an overview and links to subcategory articles - not used for RAG later
\end{enumerate}

The synthesis prompt emphasizes extracting generalizable insights, focusing on common issues and proven solutions, and organizing content into logical sections.
This consistent use of the same prompt across different synthesis strategies ensures coherent knowledge representation throughout the knowledge base.
\subsection{RAG System Architecture}

The knowledge base created by our multi-agent system serves as the foundation for a RAG system that handles supply chain queries. Our RAG implementation consists of: 1. Embedding \& Retrieval: AWS Bedrock Knowledge Base with Kendra GenAI Index, which provides semantic search capabilities across knowledge articles; 2. Response Generation: Retrieved documents are formatted into a prompt template for Claude Sonnet 3.5 v2, which generates comprehensive answers grounded in the retrieved knowledge.

\section{Experimentation}

\subsection{Datasets}

We evaluated our multi-agent approach using a real-world dataset of supply chain support tickets from Amazon internal systems. The dataset consists of tickets spanning over 1+ year of operations, split chronologically into training (80\%), validation (10\%), and test (10\%) sets. Each ticket contains ID, title, creation date, description, and comments.
This chronological splitting approach mimics real-world knowledge base creation scenarios, where future ticket data would not be available to help resolve past issues, prioritizing practical application over cross-validation generalizability. Note that in accordance with Amazon's requirements for external publications, we do not report absolute metrics on internal data and instead report all results relative to a baseline.

The tickets cover a diverse range of supply chain operational topics, including standard processes like order processing and inventory management, as well as exceptions and issues related to system access, equipment requests, and shipping concerns. The dataset exhibits several challenges typical of real-world support ticket systems: 1) Content challenges: varying levels of detail in problem descriptions, inconsistent terminology and acronyms, non-standardized resolution processes, multiple communication threads within tickets, and temporal references requiring contextual understanding; 2) Operational challenges: different support team members over time leading to variations in communication style, incomplete data with missing context, side information from parallel calls or chats not reflected in tickets, and unclosed tickets without clear resolution status. These challenges reflect the practical difficulties in knowledge extraction from operational communications in enterprise environments.

\subsection{Evaluation Setup}

We implemented and evaluated five distinct approaches to knowledge base creation:

\begin{enumerate}
    \item \textbf{Raw Ticket Indexing Baseline}: Clean raw ticket content with minimal transformation, representing a standard RAG approach over raw data
    
    \item \textbf{Single-Agent Ticket-Level Synthesis}: Individual knowledge articles created for each ticket, where each article captures generalizable insights from a single ticket

    \item \textbf{Embedding-Based Clustering Aggregation}: An approach first generates embeddings for each ticket using the Amazon titan embed model \cite{amazon2024titan}, then applies HDBSCAN \cite{campello2013density} clustering to group them into clusters representing 70.4\% of the total ticket volume, followed by generating a knowledge article for each cluster.
    
    \item \textbf{Multi-Agent Category-Driven Synthesis}: Our proposed approach using Category Discovery, Categorization, and Knowledge Synthesis agents, which create synthesized knowledge articles representing 3.4\% of the total ticket volume

    \item \textbf{Multi-Level Knowledge Integration approach}: This approach combines both Single-Agent Ticket-Level Synthesis and Multi-Agent Category-Driven Synthesis methods by retrieving the top 5 articles from each of their respective knowledge bases and merging them for generation, while our RAG system retrieves the top 10 most relevant knowledge articles for each query with other approaches.
    
\end{enumerate}

We generated test queries from the ticket titles and descriptions in the validation and test sets to simulate future internal support tickets that would need resolution. These queries represent the types of issues that internal teams would submit for resolution. To ensure fair evaluation, we generated issue descriptions that exclude any information from the ticket comments, preventing data leakage. The complete query generation prompt and process are described in Appendix \ref{query_generation_prompt}.

\subsection{Evaluation Metrics}

We evaluated the effectiveness of our approach using end-to-end RAG performance metrics:

\begin{enumerate}
    \item \textbf{Answer Helpfulness Score}: Our primary metric is a 5-point scale evaluating how helpful the generated answer would be in resolving the issue:
    \begin{itemize}
        \item 1: Not helpful at all
        \item 2: Slightly helpful
        \item 3: Moderately helpful
        \item 4: Very helpful
        \item 5: Extremely helpful
    \end{itemize}
    
    \item \textbf{Percentage of Helpful Answers}: The proportion of answers rated 4 or 5 on the helpfulness scale, representing responses that would substantially aid in issue resolution
\end{enumerate}

To ensure robust evaluation, we used Claude Sonnet 3.7 as an independent evaluator, prompted with a structured assessment framework that evaluated helpfulness based on four key dimensions: 1. Accuracy: Factual correctness based on the reference ticket; 2. Completeness: Whether the answer addresses all aspects of the query; 3. Relevance: Focus on what was actually asked; 4. Clarity: Ease of understanding. Each evaluation included the original question, the generated answer, and the reference ticket (as ground truth). The complete evaluation prompt is provided in Appendix \ref{evaluation_prompt}.
We performed 3 runs of RAG answer generation and evaluation to ensure statistical reliability, with the final metrics reflecting averaged results.

\section{Results and Discussion}

\subsection{Main Results}

Table \ref{tab:main-results} presents the performance comparison of our multi-agent approach against the baseline and alternative methods on the test set. All approaches were evaluated on both the validation and test sets using the helpfulness metrics described in Section 4.3.
 Complete validation set results are presented in Appendix \ref{validation_results} (Table \ref{tab:val-results}), which show consistent trends with the test set results, further confirming the robustness of our approach.

\begin{table*}
\caption{Performance comparison of knowledge base creation approaches on the test set. Higher values indicate better performance. Results are averaged over three runs.}
\label{tab:main-results}
\begin{tabular}{lcccc}
\toprule
\textbf{Method} & \textbf{Knowledge Articles (\%)} & \textbf{Average Helpfulness (1-5)} & \textbf{Std Dev} & \textbf{Helpful Answers (\%)} \\
\midrule
Raw Ticket Indexing Baseline & 100\% & 3.08 & 0.03 & 38.60\% \\
Single-Agent Ticket-Level Synthesis & 100\% & 3.22 & 0.02 & 40.34\% \\
Embedding-Based Clustering Aggregation & 70.4\% & 3.30 & 0.03 & 43.98\% \\
\textbf{Multi-Agent Category-Driven Synthesis} & \textbf{3.4\%} & \textbf{3.43} & \textbf{0.01} & \textbf{48.74\%} \\
Multi-Level Knowledge Integration & 103.4\% & 3.36 & 0.03 & 47.19\% \\
\bottomrule
\end{tabular}
\end{table*}

The results demonstrate a clear progression in performance across the different knowledge base creation approaches. Our Multi-Agent Category-Driven Synthesis approach achieved the best performance across both metrics on the test set, with an average helpfulness score of 3.43 out of 5 and 48.74\% of answers classified as helpful (scoring 4 or 5). This represents a significant improvement over the Raw Ticket Indexing baseline, with an 11.4\% increase in average helpfulness and a 26.3\% increase in the percentage of helpful answers.

Notably, our approach achieved these results while producing the most compact knowledge base (3.4\% of the total ticket volume), compared to the Raw Ticket Indexing baseline (100\%), Single-Agent Ticket-Level Synthesis (100\%), Embedding-Based Clustering Aggregation (70.4\%), and Multi-Level Knowledge Integration (103.4\%). This demonstrates the effectiveness of our approach in not only improving answer quality but also in consolidating knowledge into a more manageable and maintainable structure.

The performance improvement translates to practical benefits for supply chain operations. Nearly half of all queries (48.74\%) could be effectively resolved using our multi-agent knowledge base, representing a substantial opportunity for automating supply chain support and reducing manual intervention.

To understand why our multi-agent approach outperforms the alternatives, we examine the limitations of the other methods. The Raw Ticket Indexing baseline approach, while preserving all original information, suffers from several drawbacks: 1. Redundancy from similar issues appearing across multiple tickets; 2. Inclusion of irrelevant or transient information (e.g., specific dates, ticket IDs); 3. Lack of generalization from specific instances to broader patterns; 4. Absence of synthesized best practices derived from multiple related tickets.
Single-Agent Ticket-Level Synthesis improves upon the baseline by creating individual knowledge articles for each ticket, but still maintains the same volume of articles and misses opportunities for cross-ticket synthesis. The Embedding-Based Clustering Aggregation approach further improves performance by grouping similar tickets, but still results in a large number of knowledge articles that may contain overlapping information.

Our multi-agent approach addresses these limitations through its coordinated pipeline of specialized agents. The Category Discovery Agent creates a comprehensive taxonomy that organizes knowledge logically. The Ticket Categorization Agent ensures that related issues are grouped together effectively. The Knowledge Synthesis Agent then transforms these grouped tickets into comprehensive articles that capture generalizable insights across multiple related tickets.
This structure offers several advantages to the RAG system that contribute to the improved performance metrics. First, the knowledge base provides more concise, focused documents for retrieval compared to raw tickets, allowing the system to work with higher-quality context. Second, the knowledge articles contain generalized patterns rather than specific instances, improving transferability to new queries that may present variations of known issues. Third, the structured organization improves retrieval precision by ensuring that related knowledge is grouped together coherently.
By developing this high-quality knowledge base offline, we overcome many of the challenges associated with direct RAG over raw ticket data. The preprocessing eliminates redundancy and noise while enhancing the signal-to-noise ratio in the retrieved context. Moreover, the synthesis process captures connections between related issues that would be difficult to identify in real-time during query processing.

It's worth noting that the Multi-Level Knowledge Integration approach, which combines both Single-Agent Ticket-Level Synthesis and Multi-Agent Category-Driven Synthesis methods, did not outperform our Multi-Agent Category-Driven Synthesis approach. This suggests that the individual knowledge articles created for each ticket did not add more value beyond what was already captured in the synthesized knowledge articles. Furthermore, by including fewer synthesized knowledge articles in the knowledge retrieval process, the Multi-Level approach might have lost valuable information that was present in the more comprehensive Multi-Agent Category-Driven Synthesis knowledge base.

\subsection{Score Distribution Analysis}
% Figure showing score distribution comparison across methods
\begin{figure}[t]
\centering
\includegraphics[width=\linewidth]{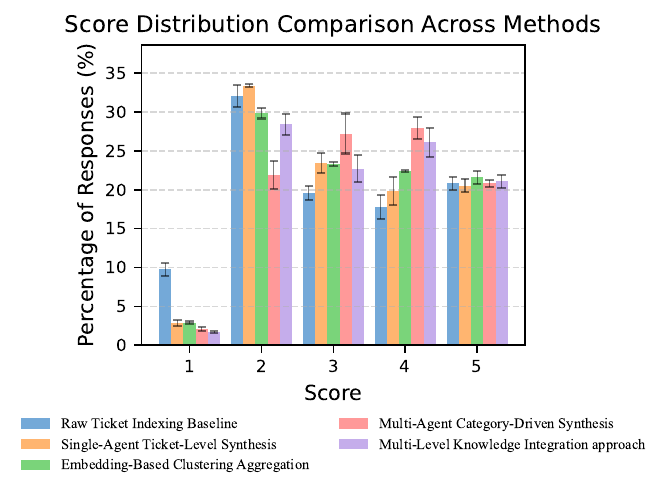}
\caption{Score Distribution Comparison Across Knowledge Creation Methods. This figure presents the percentage distribution of evaluation scores (1-5). Higher scores indicate better performance. 
% : Raw Ticket Indexing Baseline, Single-Agent Ticket-Level Synthesis, Embedding-Based Clustering Aggregation, Multi-Agent Category-Driven Synthesis, and Multi-Level Knowledge Integration. 
Each method was evaluated across three independent runs, with error bars representing the standard deviation across these runs.
% Comparison of score distributions across different methods. The scores range from 1 (Not helpful at all) to 5 (Extremely helpful). 
Multi-Agent Category-Driven Synthesis shows a substantial reduction in scores 1 and 2 responses and an increase in higher scores 3 and 4 compared to the Raw Ticket Indexing baseline, indicating improved response quality.}
\label{fig:score-distribution}
\end{figure}
% Figure: Score distribution comparison
To gain deeper insights into the nature of our approach's improvements, we performed a detailed analysis of the score distribution patterns. This analysis reveals not just aggregate improvements but also how our multi-agent system changes the quality profile of the generated responses.

We performed a detailed analysis of the score distribution to better understand the improvements offered by our approach. From the score distribution data of the Raw Ticket Indexing baseline and Multi-Agent Category-Driven Synthesis, we observed significant shifts in the helpfulness ratings. The most striking improvement is the dramatic reduction in score=1 responses (completely unhelpful), which decreased by 77.4\% compared to the Raw Ticket Indexing baseline. Meanwhile, there were substantial increases in both moderately helpful (score=3, +43.8\%) and very helpful (score=4, +66.1\%) responses. The overall distribution shifted notably toward higher scores.

This distribution shift is meaningful in practical applications. By substantially reducing the proportion of unhelpful responses (scores 1-2) while increasing helpful ones (scores 4-5), our multi-agent approach provides more reliable support for supply chain queries. The shift toward more moderate and helpful responses suggests that our knowledge base more effectively captures the generalizable patterns needed to address a wide range of supply chain issues.

\subsection{Statistical Significance}

To verify the statistical significance of our results, we conducted a Welch's t-test comparing the mean helpfulness scores of our Multi-Agent Category-Driven Synthesis approach against the Raw Ticket Indexing baseline. Statistical analysis of the results across the three evaluation runs showed that the improvement of our approach over the baseline (3.43 vs 3.08) is highly statistically significant (p < 0.001). 

The consistency across multiple evaluation runs, evidenced by the low standard deviation (0.01) for Multi-Agent Category-Driven Synthesis compared to other approaches, further strengthens the reliability of these findings. The magnitude of improvement, particularly in reducing unhelpful responses (scores 1-2) by 43.2\% while increasing helpful responses by 24.8\%, demonstrates substantial practical significance in the improvements provided by our approach.

These statistical findings confirm that the performance advantages of our multi-agent knowledge base creation approach are robust and unlikely to be due to chance or evaluation variability. The combination of statistical significance with meaningful practical improvements in answer helpfulness provides strong evidence for the effectiveness of our approach in real-world supply chain support scenarios.

\subsection{Limitations and Future Work}

While our approach shows promising results, several limitations should be noted: 1. Domain-specific terminology and acronyms remain challenging for LLMs to process without additional context; 2. The effectiveness of the approach depends on the quality and consistency of the initial ticket data; 3. The current implementation does not handle multimodal information (images, attachments) that may be crucial for some tickets; 4. Knowledge base maintenance and updates over time require further investigation.

The first limitation reflects a broader challenge in specialized domains like supply chain management—terminology is often company-specific or industry-specific and may evolve over time. Future work could address this through domain adaptation techniques or specialized training.

The dependence on initial ticket quality presents both a limitation and an opportunity. While inconsistent or incomplete tickets may lead to lower-quality knowledge articles, the structured approach we've developed could potentially be used to improve the ticket creation process itself, prompting users for more complete and consistent information.

The inability to handle multimodal information is particularly relevant in supply chain contexts where visual information (e.g., damaged products, equipment states) often plays a crucial role in problem diagnosis and resolution. Extending our approach to incorporate multimodal data represents an important avenue for future research.

Finally, the question of knowledge base maintenance highlights the need for continuous learning and updating mechanisms. As supply chain processes and systems evolve, the knowledge base must adapt accordingly. Developing efficient update processes that don't require rebuilding the entire knowledge base is an important area for future work.

Despite these limitations, our results demonstrate that the multi-agent approach offers significant improvements over traditional RAG implementations and simpler single-agent and embedding-based clustering knowledge base creation methods. The substantial performance gains achieved with our approach provide a strong foundation for future refinements to address these limitations.

% Future work will focus on extending the system to handle multimodal inputs (e.g., images, diagrams, and spreadsheets attached to tickets) and the evolution of knowledge base, and exploring the use of the extracted knowledge for automated decision support and proactive recommendation.

\section{Conclusion}

In this paper, we presented a simple but effective LLM-based multi-agent approach for knowledge base creation from ticket systems in supply chain contexts. Our system transforms unstructured communications into structured, reusable, and actionable knowledge base through specialized agents for category discovery, categorization, and knowledge synthesis. Our experimental results demonstrate that plugging this prebuilt knowledge base into a RAG system significantly improves the helpfulness of generated answers and aids in resolving about 50\% of future tickets. These findings suggest that our approach can significantly improve knowledge reuse and issue resolution in supply chain operations by capturing and formalizing the institutional knowledge that is typically lost in communication channels.

\begin{acks}
Thanks to Tom Emerton for sharing the ticket data received and addressed by their team. Also, thanks to Zach Roberts, Sheilla Wambua, and leaders at Amazon Operational Technology Solutions team for their support to this work. We also thank the anonymous
reviewers for their valuable feedback.
\end{acks}

\bibliographystyle{ACM-Reference-Format}
\bibliography{reference}

\appendix
\section{Summary Table of Related Work}\label{related_work_table}
% Table. Comparison of recent LLM/RAG/agentic systems for extracting structured knowledge from enterprise data. Each row summarizes a representative work's task, data, methods, and evaluation metrics.
Table \ref{tab:llm-research} provides a comprehensive comparison of recent research on LLM-based, RAG, and multi-agent systems for extracting structured knowledge from enterprise data.
\begin{table*}
\caption{Comparison of recent LLM/RAG/agentic systems for extracting structured knowledge from enterprise data (LLM = large language model; RAG = retrieval-augmented generation.)}
\label{tab:llm-research}
\begin{tabular}{p{2.2cm}|p{2.2cm}|p{2.2cm}|p{4cm}|p{4cm}}
\toprule
\textbf{Work (Year)} & \textbf{Task / Domain} & \textbf{Data} & \textbf{Methods} & \textbf{Evaluation} \\
\midrule
Anderson et al. (2025) \cite{anderson_design_2025} & Analytics QA on documents & Unstructured reports (e.g. NTSB) & LLM-based query planner + document operations & Query-answer accuracy vs. RAG \\
\midrule
Kumar et al. (2025) \cite{kumar_llm-powered_2025} & Enterprise knowledge graph & Internal data (email/chat/logs) & LLM entity/relation extraction $\rightarrow$ KG & Qualitative task improvements (expertise discovery, etc.) \\
\midrule
Wang et al. (2024) \cite{wang_evopat_2024} & Patent summarization & Patent corpus (global) & Multi-LLM agents + RAG search & ROUGE and human judgments; outperforms GPT-4 \\
\midrule
AlMahri et al. (2024) \cite{almahri_enhancing_2024} & Supply-chain mapping & Public reports/news (EV supply) & Zero-shot LLM NER/RE $\rightarrow$ KG & NER/RE accuracy; visibility beyond Tier-1 suppliers \\
\midrule
Houamegni \& Gedikli (2025) \cite{houamegni_evaluating_2025} & News summarization (SC) & News articles (supply chain) & LLM summarization (zero/few-shot) & ROUGE/BLEU, BERTScore, user study (risk identification) \\
\midrule
Edge et al. (2025) \cite{edge_local_2025} & Query-focused summarization & Large document sets ($\sim$10$^6$ token) & Graph RAG: LLM builds entity graph + cluster summaries & Answer diversity/coverage vs. vanilla RAG \\
\midrule
LlamaIndex ADW (2025) & Document workflows & Enterprise docs (PDF/contracts) & Agent pipeline (Parse$\rightarrow$Retrieve$\rightarrow$Reason$\rightarrow$Act) & Demo (contract clause analysis); no benchmark \\
\midrule
Pathway (2024) & Multi-agent RAG & Legal/financial long docs & Supervisor + RAG agent + Reasoning agent & Improved retrieval quality (MRR) and answer accuracy \\
\midrule
Crawford et al. (2024) \cite{crawford_bmw_2024} & Multi-agent automation & Synthetic enterprise tasks & Planner/Coordinator + specialized agents (search, editor, tester) & Case-study workflows (QA, editing) \\
\bottomrule
\end{tabular}
\end{table*}

\section{Agent Prompts}\label{agent_prompts}

This appendix provides the key prompts used for each agent in our multi-agent knowledge base creation system. These prompts were engineered to guide Claude Sonnet 3.7 in performing specific tasks within the knowledge extraction pipeline.

\subsection{Category Discovery Agent Prompt}\label{category_discovery_prompt}

The Category Discovery Agent analyzes ticket data to identify distinct knowledge domains and create a taxonomy of categories:

\begin{lstlisting}[style=promptstyle]
You are analyzing ticket data from a supply chain management system. Your task is to discover knowledge categories based STRICTLY on the provided sample tickets.

# Sample Tickets
```
{sample_tickets}
```

# Task
Create a taxonomy of knowledge categories based ONLY on these sample tickets.

For each category:
1. Provide a clear, concise name to capture the essence of the issue type (5 words or less)
2. Write a brief description of what this category encompasses (50 words or less)
3. List identifying patterns or keywords (maximum 15 per category)

# Important Guidelines
1. Focus ONLY on categories that are ACTUALLY REPRESENTED in the sample tickets
2. Categories should be based on the nature of the problem, not just surface details
3. Categories should be distinct from each other with minimal overlap
4. The number of categories should reflect the diversity in the sample - DO NOT create more categories than justified by the samples
5. Be extremely concise with category names and description, and use short keywords for identifying patterns
6. DO NOT create categories for general organizational content that isn't an actual problem
7. DO NOT use your general knowledge about supply chain systems to invent categories - rely ONLY on what's in the data

# Output Format
Return a JSON structure:

{
  "categories": [
    {
      "name": "Short Category Name",
      "description": "Brief description",
      "identifying_patterns": ["pattern1", "pattern2", "pattern3"]
    }
  ]
}

Ensure your JSON is properly formatted and valid. Be extremely concise with all text to minimize token usage.
\end{lstlisting}

\subsection{Category Merge Prompt}\label{category_merge_prompt}

This prompt is used for merging multiple category sets discovered from different batches:

\begin{lstlisting}[style=promptstyle]
You are analyzing multiple sets of knowledge categories from a supply chain management system. Your task is to merge these category sets into one comprehensive, non-redundant taxonomy.

# Category Sets
```
{category_sets_json}
```

# Task
Merge these category sets into a single comprehensive taxonomy. Each category set was derived from different batches of ticket data, so they may contain:
1. Identical categories that appear in multiple sets
2. Similar categories with slight variations in name, description, or identifying patterns
3. Unique categories that only appear in one set

For the merged taxonomy:
1. Consolidate identical or highly similar categories
2. For similar categories, combine the best elements of each description and merge identifying patterns
3. Preserve unique categories that represent distinct knowledge areas
4. Keep descriptions brief (50 words maximum)
5. Limit to maximum 15 identifying patterns per category

# Important Guidelines
1. Focus on semantic similarity, not just text matching
2. When merging similar categories, choose the clearest name and most comprehensive description
3. Combine identifying patterns from similar categories but avoid redundancy
4. Ensure the final taxonomy is comprehensive with minimal redundancy
5. Be extremely concise with category names and descriptions

# Output Format
Return a JSON structure with the merged categories:

{
  "categories": [
    {
      "name": "Short Category Name",
      "description": "Brief description",
      "identifying_patterns": ["pattern1", "pattern2", "pattern3"]
    }
  ],
  "merge_summary": "Brief description of how the merging was done"
}

Ensure your JSON is properly formatted and valid. Be extremely concise with all text to minimize token usage.
\end{lstlisting}

\subsection{Subcategory Discovery Agent Prompt}\label{subcategory_discovery_prompt}

For large categories (exceeding a configurable threshold), the system discovers more specific subcategories:

\begin{lstlisting}[style=promptstyle]
You are analyzing a set of tickets from a supply chain management system that have already been categorized into a main category. Your task is to discover SUBCATEGORIES within this main category.

# Main Category Information
Name: {parent_category_name}
Description: {parent_category_description}

# Sample Tickets from this Category
```
{sample_tickets}
```

# Task
Create a taxonomy of SUBCATEGORIES for these tickets. These tickets all belong to the same main category, but need to be further organized into more specific subcategories.

For each subcategory:
1. Provide a clear, concise name to capture the specific issue type (5 words or less)
2. Write a brief description of what this subcategory encompasses (50 words or less)
3. List identifying patterns or keywords (maximum 10 per subcategory)

# Important Guidelines
1. Focus ONLY on subcategories that are ACTUALLY REPRESENTED in the sample tickets
2. The subcategories should be distinct from each other with minimal overlap
3. The number of subcategories should reflect the diversity in the sample - DO NOT create more subcategories than justified by the samples
4. The subcategory names should clearly relate to the parent category but be more specific
5. Be extremely concise with subcategory names and descriptions
6. DO NOT use your general knowledge to invent subcategories - rely ONLY on what's in the data

# Output Format
Return a JSON structure:

{
  "subcategories": [
    {
      "name": "Short Subcategory Name",
      "description": "Brief description",
      "identifying_patterns": ["pattern1", "pattern2", "pattern3"],
      "parent_category": "{parent_category_name}"
    }
  ]
}

Ensure your JSON is properly formatted and valid. Be extremely concise with all text to minimize token usage.
\end{lstlisting}

\subsection{Ticket Categorization Agent Prompt}\label{ticket_categorization_prompt}

The Categorization Agent assigns each ticket to one or more categories:

\begin{lstlisting}[style=promptstyle]
You are categorizing a supply chain ticket into predefined knowledge categories.

# Ticket Information
Title: {title}
Description: {description}

# Available Categories
{categories}

# Task
Assign this ticket to the most appropriate category from the list.
If the ticket clearly fits multiple categories, you may assign it to up to 2 categories.

# Output Format
Return a JSON structure:

{
  "assignments": [
    {
      "category": "Category Name",
      "reasoning": "Brief explanation of why this category fits"
    }
  ]
}

If no categories are clearly applicable, return an empty assignments array.
\end{lstlisting}

\subsection{Subcategory Categorization Agent Prompt}\label{subcategory_categorization_prompt}

After discovering subcategories, tickets are assigned to the appropriate subcategory using this specialized prompt:

\begin{lstlisting}[style=promptstyle]
You are categorizing a ticket into predefined subcategories within a main category.

# Ticket Information
Title: {title}
Description: {description}

# Main Category
{parent_category_name}: {parent_category_description}

# Available Subcategories
{subcategories}

# Task
Assign this ticket to the most appropriate subcategory from the list.

# Output Format
Return a JSON structure:

{
  "assignments": [
    {
      "subcategory": "Subcategory Name",
      "reasoning": "Brief explanation of why this subcategory fits"
    }
  ]
}

If no subcategories are clearly applicable, return an empty assignments array.
\end{lstlisting}

\subsection{Knowledge Synthesis Agent Prompt}\label{knowledge_synthesis_prompt}

The Knowledge Synthesis Agent transforms groups of categorized tickets into comprehensive knowledge articles:

\begin{lstlisting}[style=promptstyle]
You are synthesizing knowledge from supply chain tickets to create a concise, factual knowledge base article specifically for users who create tickets in this system.

# Category Information
Name: {category_name}
Description: {category_description}

# Tickets in this Category
{ticket_data}

# Task
Create a CONCISE knowledge article that contains ONLY information directly supported by the ticket data. 

IMPORTANT REQUIREMENTS:
1. Use ONLY information explicitly mentioned in the ticket data
2. DO NOT expand acronyms unless they are expanded in the tickets themselves
3. DO NOT make up definitions for systems if not provided in the data
4. DO NOT invent processes or best practices not mentioned in tickets
5. Keep the article SHORT and FOCUSED - aim for 50% less content than you might typically write
6. Write in a direct style addressing ticket creators

Focus on:
1. Common issues seen in these tickets (briefly)
2. Actual solutions that worked (from ticket resolutions)
3. Minimal, specific advice based only on ticket content

# Output Format
Your response should be a concise markdown document with:

1. Title: A brief descriptive title
2. Common Issues: 2-3 bullet points of the main issues (be brief)
3. Tips for Resolution: Specific advice based ONLY on what worked in the tickets
4. Resources: Only mention systems/links that appear in the tickets

Total length should be no more than 400-500 words maximum.
\end{lstlisting}

\subsection{Knowledge Merge Prompt}\label{knowledge_merge_prompt}

This prompt is used for merging multiple knowledge articles on the same topic:

\begin{lstlisting}[style=promptstyle]
You are merging multiple knowledge articles on the same topic into a single comprehensive article.

# Category Information
Name: {category_name}
Description: {category_description}

# Knowledge Articles to Merge
{articles_to_merge}

# Task
Create a single, coherent knowledge article that combines insights from all the provided articles.

IMPORTANT REQUIREMENTS:
1. ORGANIZE information logically into sections (Common Issues, Tips for Resolution, Resources)
2. REMOVE redundancy - multiple articles may cover the same points
3. PRIORITIZE information that appears in multiple articles
4. INCLUDE unique insights from individual articles if they add value
5. MAINTAIN conciseness - focus on the most valuable information
6. USE the same level of specificity as the input articles
7. DO NOT introduce new information not present in the source articles
8. DO NOT expand acronyms unless they were expanded in the source articles

# Output Format
Your response should be a single markdown document with:

1. Title: A clear descriptive title related to the category
2. Common Issues: Consolidated list of key issues (brief bullet points)
3. Tips for Resolution: Specific advice based on the source articles
4. Resources: Systems/links that appear in the source articles

Total length should be no more than 400-500 words maximum.
\end{lstlisting}

\section{Evaluation Prompt}\label{evaluation_prompt}

This section presents the evaluation prompt used to assess the helpfulness of RAG-generated answers. The evaluations were conducted using Claude Sonnet 3.7 with the following prompt:

\begin{lstlisting}[style=promptstyle]
You are an expert evaluator assessing the helpfulness of answers to customer service questions.

# Question
{question}

# Answer to Evaluate
{answer}

# Original Ticket (Reference)
{ticket_content}

# Evaluation Task
Please evaluate if the answer is helpful for the given question, considering the original ticket as the source of truth.

1. Accuracy: Does the answer provide factually correct information based on the ticket?
2. Completeness: Does the answer address all aspects of the question?
3. Relevance: Does the answer focus on what was actually asked?
4. Clarity: Is the answer easy to understand?

# Output Format
Provide your evaluation in the following format:
1. Helpfulness Score (1-5, where 1 is not helpful at all and 5 is extremely helpful)
2. Reasoning: Brief explanation for your score
3. Missing Information: Any critical information from the ticket that should have been included
4. Improvement Suggestions: How the answer could be made more helpful
\end{lstlisting}

The evaluation process assessed each answer on a 5-point helpfulness scale:
\begin{itemize}
    \item 1: Not helpful at all
    \item 2: Slightly helpful
    \item 3: Moderately helpful
    \item 4: Very helpful
    \item 5: Extremely helpful
\end{itemize}

The evaluation focused on four key dimensions: accuracy (factual correctness based on the reference ticket), completeness (addressing all aspects of the query), relevance (focusing on what was actually asked), and clarity (ease of understanding). This comprehensive evaluation framework ensured a rigorous and consistent assessment of RAG system performance across different knowledge base creation approaches.

\section{Query Generation Process}\label{query_generation_prompt}

For the evaluation described in Section 4.2, we generated test queries from ticket titles and descriptions using Claude Sonnet 3.7 with the following prompt:

\begin{lstlisting}[style=promptstyle]
You are analyzing a support ticket from a supply chain management system. Your task is to generate a concise query that represents the core issue or question in this ticket.

# Ticket Information
Title: {title}
Description: {description}

# Task
Create a concise query (1 sentence) that captures the main issue or request from this ticket. The query should:
1. Be phrased as a clear, specific question or problem statement
2. Include relevant context (site codes, order numbers, etc.) if they are critical to understanding the issue
3. Focus on what the requester needs help with
4. Be concise but complete (typically 5-15 words)

# Output Format
Return ONLY the query text with no prefixes, explanations, or surrounding quotes.
\end{lstlisting}

This process transformed ticket information into natural language queries that simulate how users would typically ask for assistance with similar supply chain issues, ensuring that our evaluation reflected realistic usage scenarios. The implementation included features for parallel processing, rate limiting, and caching to enable efficient generation of queries for all validation and test set tickets.

\section{Validation Set Results}\label{validation_results}

This section presents the results of our evaluation on the validation set, complementing the test set results presented in Section 5.1. Table \ref{tab:val-results} shows the performance comparison of our multi-agent approach against the baseline and alternative methods on the validation set.

\begin{table*}
\caption{Performance comparison of knowledge base creation approaches on the validation set. Higher values indicate better performance. Results are averaged over three runs.}
\label{tab:val-results}
\begin{tabular}{lcccc}
\toprule
\textbf{Method} & \textbf{Knowledge Articles (\%)} & \textbf{Average Helpfulness (1-5)} & \textbf{Std Dev} & \textbf{Helpful Answers (\%)} \\
\midrule
Raw Ticket Indexing Baseline & 100\% & 3.00 & 0.02 & 41.27\% \\
Single-Agent Ticket-Level Synthesis & 100\% & 3.14 & 0.01 & 41.74\% \\
Embedding-Based Clustering Aggregation & 70.4\% & 3.27 & 0.00 & 45.71\% \\
\textbf{Multi-Agent Category-Driven Synthesis} & \textbf{3.4\%} & \textbf{3.31} & \textbf{0.02} & \textbf{44.94\%} \\
Multi-Level Knowledge Integration & 103.4\% & 3.30 & 0.03 & 47.13\% \\
\bottomrule
\end{tabular}
\end{table*}

The validation set results follow similar trends to those observed on the test set. Our Multi-Agent Category-Driven Synthesis approach achieved strong performance with an average helpfulness score of 3.31 out of 5 and 44.94\% of answers classified as helpful (scoring 4 or 5). This represents a 10.3\% improvement in average helpfulness over the Raw Ticket Indexing baseline.

Interestingly, on the validation set, the Multi-Level Knowledge Integration approach achieved the highest percentage of helpful answers at 47.13\%, slightly outperforming the Multi-Agent Category-Driven Synthesis approach in this metric. However, our Multi-Agent Category-Driven Synthesis approach maintained the highest average helpfulness score and achieved these results with the most compact knowledge base (3.4\% of the total ticket volume), demonstrating its efficiency in knowledge consolidation.

The consistent performance across both validation and test sets strengthens the reliability of our findings and confirms that the improvements provided by our multi-agent approach are robust across different data samples.

\end{document}